\pdfoutput=1

\documentclass[11pt]{article}

\usepackage{authblk}

\usepackage{acl}

\usepackage{times}
\usepackage{latexsym}

\usepackage[T1]{fontenc}

\usepackage[utf8]{inputenc}

\usepackage{microtype}

\usepackage{amsmath}
\usepackage{amssymb}
\usepackage{booktabs}
\usepackage{subfig}
\usepackage{graphicx}
\usepackage{enumerate}
\usepackage[noabbrev]{cleveref}

\newcommand{\mask}{\underline{\hspace{1.66em}}}
\newcommand{\lot}{\text{Leap-of-Thought}}
\newcommand{\conceptnet}{\textsc{ConceptNet}}

\title{Does Pre-training Induce Systematic Inference? \\ How Masked Language Models Acquire Commonsense Knowledge}

\author[1]{Ian Porada}
\author[2]{Alessandro Sordoni}
\author[1]{Jackie Chi Kit Cheung}
\affil[1]{Mila, McGill University}
\affil[ ]{{\tt \{ian.porada@mail, jcheung@cs\}.mcgill.ca}}
\affil[2]{Microsoft Research Montr\'eal}
\affil[ ]{{\tt alsordon@microsoft.com}}

\begin{document}

\maketitle

\begin{abstract}
Transformer models pre-trained with a masked-language-modeling objective (e.g., BERT) encode commonsense knowledge as evidenced by behavioral probes; however, the extent to which this knowledge is acquired by systematic inference over the semantics of the pre-training corpora is an open question. To answer this question, we selectively inject verbalized knowledge into the minibatches of a BERT model during pre-training and evaluate how well the model generalizes to supported inferences. We find generalization does not improve over the course of pre-training, suggesting that commonsense knowledge is acquired from surface-level, co-occurrence patterns rather than induced, systematic reasoning.
\end{abstract}

\section{Introduction}
\label{sec:introduction}

Pre-trained Transformers, such as BERT, encode knowledge about the world \citep{petroni-etal-2019-language, zhou-etal-2020-evaluating}; e.g., BERT assigns relatively high probability to ``fly'' appearing in the context ``robins can \mask{} .'' In this work, we investigate whether such knowledge is acquired during pre-training through systematic inference over the semantics of the pre-training corpora; e.g., can models systematically infer ``robins can fly'' from the premises ``birds can fly'' and ``robins are birds?''

Resolving \emph{how} models acquire commonsense knowledge has important implications. If models learn to make systematic inferences through pre-training, then scaling up pre-training is a promising direction for commonsense knowledge acquisition. If, instead, models only ever generalize based on surface-level patterns, then the majority of commonsense knowledge, which is only supported implicitly, will never be acquired \citep{10.1145/2509558.2509563, forbes-choi-2017-verb}.

On the one hand, there is cursory evidence that pre-training might induce the ability to systematically reason about the world. When fine-tuned on supervised training sets, pre-trained models can classify valid inferences better than strong baselines \citep{clark-etal-2020-ruletaker, talmor-2020-leap}; and, in zero-shot evaluations, pre-trained models perform relatively well on reasoning tasks that \textit{may} require systematic reasoning, such as number comparison \citep{talmor-etal-2020-olmpics} and Winograd schemas \cite{10.1145/3474381}.

On the other hand, existing works have argued that pre-training does not generalize by systematic inference over semantics on the basis of theoretical or synthetic results \cite{bender-koller-2020-climbing, 10.1162/tacl_a_00412, traylor-etal-2021-mean}. Referring to physical commonsense knowledge acquired by BERT, \citet{forbes2019neural} conclude that ``neural language representations still only learn associations that are explicitly written down.''

Our main contribution is a direct evaluation of the training dynamics of BERT's reasoning ability. We inject verbalized knowledge, such as ``robins can fly'' (where the masked token is the predicate, e.g., ``fly''), into the minibatches of BERT throughout pre-training. We then consider how well BERT generalizes to supported inferences; e.g., how does the likelihood of ``robins are \mask{}'' $\rightarrow{}$ ``birds'' change?

We find generalization does not improve over the majority of pre-training which supports the hypothesis that commonsense knowledge is not acquired by systematic inference. Rather, our findings suggest knowledge is acquired from surface-level, co-occurrence patterns.

\section{Related Work}
\label{sec:related-work}

Commonsense knowledge acquisition is a longstanding challenge in natural language processing \citep{charniak-1973-jack, hwang-etal-2020-atomic, zhang2021aser}, and current approaches rely on knowledge acquired by pre-training Transformer language models \citep{bosselut-etal-2019-comet, ijcai2020-554, west2021symbolic}. The commonsense reasoning ability of pre-trained models has been evaluated using behavioral probes \citep{Ettinger-tacl_a_00298, misra2021typicality, he-etal-2021-winologic} and downstream, fine-tuned evaluations \citep{banerjee2021commonsense, zhou-etal-2021-rica}. While these works only consider the ability of a fully pre-trained model, we evaluate how knowledge acquisition develops throughout pre-training.

When fine-tuned on supervised datasets, pre-trained models can learn to make systematic inferences to some extent \citep{clark-etal-2020-ruletaker, tafjord-etal-2021-proofwriter, NEURIPS2020_fc84ad56, shaw-etal-2021-compositional, li-etal-2021-implicit}. Here, by \emph{systematic inferences}, we refer to the ability to learn general rules and apply them in novel settings \citep{FODOR19883, Lake2018GeneralizationWS, bahdanau2018systematic} as opposed to learning only some particular instances of the rule.

As in our experiments, recent work has also considered the training dynamics of pre-trained models \citep{NEURIPS2020_1457c0d6, Kaplan2020ScalingLF}. \citet{liu-etal-2021-probing-across} specifically consider zero-shot performance of RoBERTa on the oLMpics reasoning tasks \citep{talmor-etal-2020-olmpics}, but find the knowledge studied is never learned. In contrast, we explore \emph{how} learned knowledge is acquired.

Close in spirit to our work, \citet{kassner-etal-2020-pretrained} pre-train a masked language model on a synthetic dataset to isolate reasoning ability. \citet{wei-etal-2021-frequency} also intervene on BERT's pre-training data in a syntactic evaluation and conclude that subject-verb agreement is inferred from rules to some extent.

Finally, \citet{de-cao-etal-2021-editing} explore how knowledge encoded in BERT is affected by gradient updates when fine-tuning on a downstream task. \citet{hase2021language} build on this work and explore how updates on premises affect supported knowledge. Our work is unique, however, in that we focus on pre-training itself which we contrast with fine-tuning evaluations (\S\ref{sec:finetuning}).

\section{Method}
\label{sec:method}

The purpose of our evaluation is to answer the question: \emph{does BERT systematically infer commonsense knowledge from premises present in the pre-training corpus?}

We focus on commonsense knowledge that BERT is known to encode, namely simple entity properties such as those annotated in \conceptnet{} \citep{conceptnet}. This knowledge can be represented abstractly as (\texttt{subject}, \texttt{predicate}, \texttt{object}) triples. We verify BERT's encoding of knowledge by the ability to predict the \texttt{object} conditioned on a verbalization of the knowledge containing the \texttt{subject} and \texttt{predicate}; e.g., for (\texttt{robin}, \texttt{capable-of}, \texttt{fly}), we evaluate the ability to predict ``fly'' appearing in the context ``robins can \mask{} .''

\begin{table}[t]
\begin{tabular}{c@{\hskip 2em}c}
\toprule
Type & Example \\
\midrule
Super-statement & \textit{A bird can \mask{} . $\rightarrow{}$ fly} \\
Sub-statement & \textit{A robin can \mask{} . $\rightarrow{}$ fly} \\
Class Relation & \textit{A robin is a \mask{} . $\rightarrow{}$ bird} \\
\bottomrule
\end{tabular}
    \caption{Illustrative example of the three knowledge types as masked-token prediction.}
    \label{table:knowledge-types}
\end{table}

This knowledge may be supported by simple co-occurrence patterns (such as ``robin'' and ``fly'' having high co-occurrence), but we are interested in the extent to which knowledge might also be supported by induced, systematic inference. We focus on the inference of \emph{downward monotonicity} (\emph{A is-a B $\land$ B has-property C $\vDash$ A has-property C}). We refer to the hypernym property (\emph{B has-property C}) as the super-statement, the hyponym property (\emph{A has-property C}) as the sub-statement, and the hypernymy relation (\emph{A is-a B}) simply as the class-relation (Table \ref{table:knowledge-types}).

We can then evaluate, for example, whether ``robins can fly'' is influenced by the inference ``robins are birds'' $\land$ ``birds can fly'' $\vDash$ ``robins can fly.'' For this evaluation, we inject a supporting premise into a pre-training minibatch (i.e., we replace one of the sentences in the minibatch with the premise) and then evaluate knowledge of the supported inference after updating on the premise.

We run this evaluation across time, evaluating BERT at several checkpoints throughout pre-training. If pre-training induces the ability to systematically make the downward monotonicity inference, one would expect that generalization from premise to inference will improve during pre-training.

\subsection{Metrics}
\label{sec:metrics}

Let $\theta_i$ be the parameterization of BERT at pre-training iteration $i$, and let be $w = (x, y, z)$ be a knowledge instance where $x$ is the corresponding super-statement, $y$ the sub-statement, and $z$ the class-relation.

We take $x$ to be a logical premise. Let $\theta_i^x$ be $\theta_i$ after one gradient update on a minibatch containing $x$. For a hypothesis $h$ (which is a possible inference and could be $x$, $y$, or $z$), we consider:

\begin{enumerate}[{(1)}]
    \item Prior log-probability: $\log p(h | \theta_i)$
    \item Posterior log-probability:  $\log p(h | \theta_i^x)$
    \item PMI: $\log p(h | \theta_i^x) - \log p(h | \theta_i)$
\end{enumerate}

Intuitively, (1) describes the model's prior knowledge of $h$ at step $i$, and (3) describes how a pre-training update on $x$ affects the knowledge of $h$. We also consider standard informational retrieval metrics such as mean reciprocal rank (MRR).

\section{Experiments}
\label{sec:experiments}

\subsection{Inference Dataset}
\label{sec:dataset}

We evaluate on the \lot{} dataset presented by \citet{talmor-2020-leap}. This is a dataset of 30k true or false downward-monotonic inferences. The hypernymy relations are taken from WordNet \citep{wordnet}, and the properties are taken from both WordNet and \conceptnet{} \citep{conceptnet}.

We reformulate this supervised, classification dataset as a zero-shot, cloze-style task. First, we filter the dataset by removing examples where one type of knowledge is withheld. Then, we filter out the randomly-generated negative examples (e.g. ``a car cannot fly''), and those where the predicate is longer than one word-piece. (This last step follows the procedure of the LAMA evaluation \citep{petroni-etal-2019-language} and allows us to evaluate BERT in a zero-shot setting.) The filtered dataset consists of 711 examples. For each example, the knowledge is converted into a cloze task by masking the predicate.

To evaluate relative performance, we also generate a control for each example by randomly sampling a WordNet sibling of the super-statement hypernym as a pseudo-negative; e.g., controls are of the form ``A robin is a \mask{} .'' $\rightarrow{}$ ``fish.'' We use a property of the control entity as a control predicate for the super and sub-statements.

\subsection{Model}
\label{sec:model}

We evaluate the training dynamics of a BERT-base model \citep{devlin-etal-2019-bert} with whole-word masking and sentence-order prediction \citep{lan-etal-2020-albert}. We pre-train the model for 1 million steps on a concatenation of English Wikipedia and the Toronto Book Corpus \citep{zhu-etal-2015-aligning} as released by Huggingface datasets \citep{lhoest-etal-2021-datasets}. The training corpora are sentence tokenized using NLTK Punkt tokenizer \citep{bird-loper-2004-nltk}, and these sentences used as training sequences instead of random spans of text as used in the original BERT.

\begin{figure}[t]
\includegraphics[width=8cm]{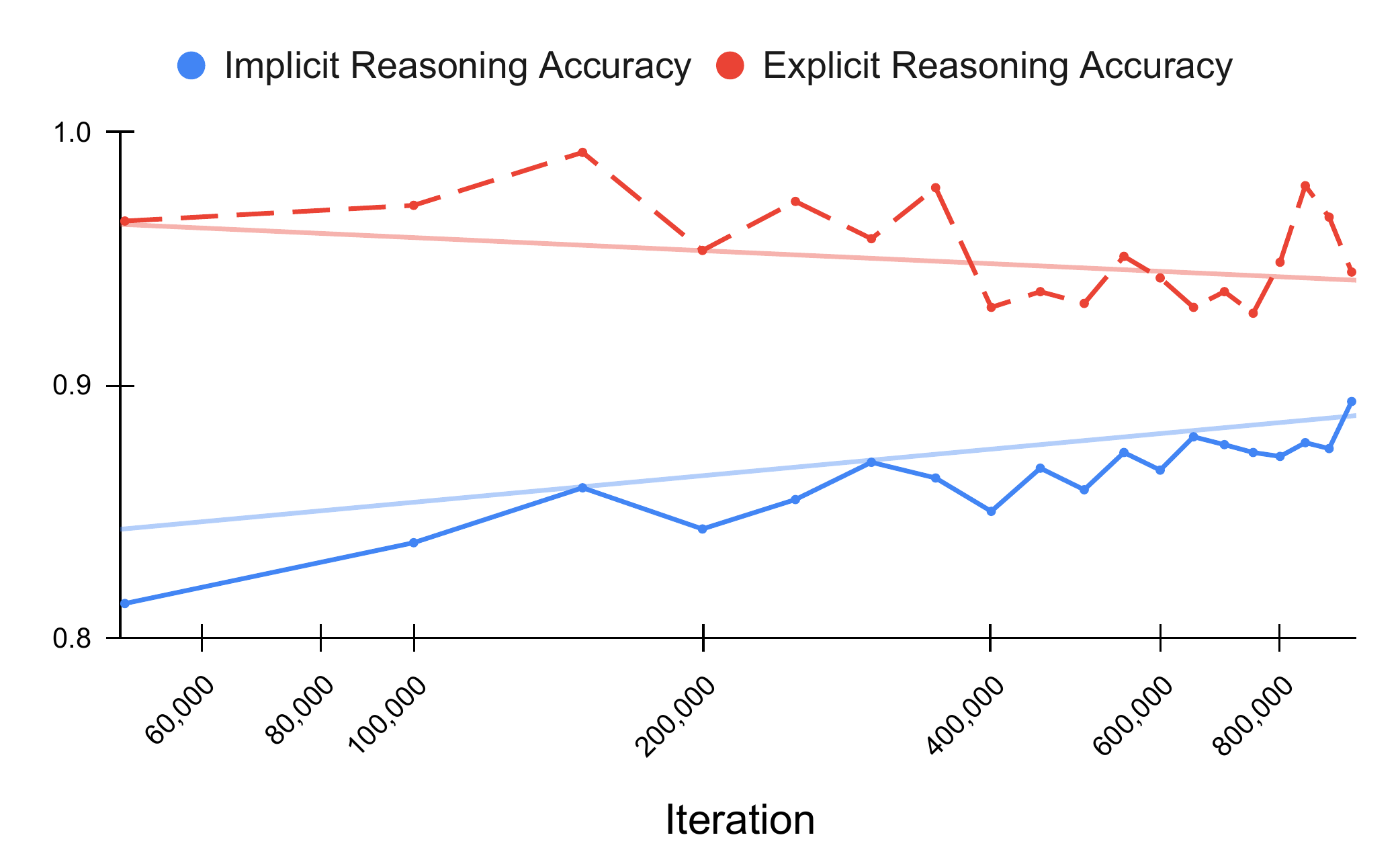}
\caption{Accuracy on \citet{talmor-2020-leap}'s original \lot{} evaluation across pre-training iterations (from 50k to 1M).}
\label{fig:motivation}
\end{figure}

Closely following the original hyperparameters of BERT, we use a batch size of 256, sequence length of 128, and train for 1 million steps. We linearly warmup the learning rate to 1e-4 over the first 10,000 steps of pre-training, and then linearly decay the learning rate. Additional hyperparameters can be found in Appendix \ref{sec:bert-hyperparameters}. Our code builds on the Huggingface Transformers \citep{wolf-etal-2020-transformers} and MegatronLM \citep{shoeybi2019megatronlm} implementations of BERT. For evaluating training dynamics, we save a checkpoint of the model and optimizer throughout pre-training.

At each pre-training checkpoint, we perform the pre-training intervention. Specifically, we inject 20 random super-statements into a minibatch and perform one gradient update on this minibatch using the saved optimizer and a constant learning rate of 1e-4 (to control for the effects of the learning rate scheduler). We then evaluate change in likelihood of the inferences supported by the injected super-statements. We continue this evaluation 200 times so that each of the 711 Leap-of-Thought examples has been evaluated in five separate minibatches. We evaluate 20 pre-training checkpoints with this procedure.

\begin{figure*}[t]%
    \centering
    \subfloat[\centering Prior Log-Prob.]{{\includegraphics[width=5cm]{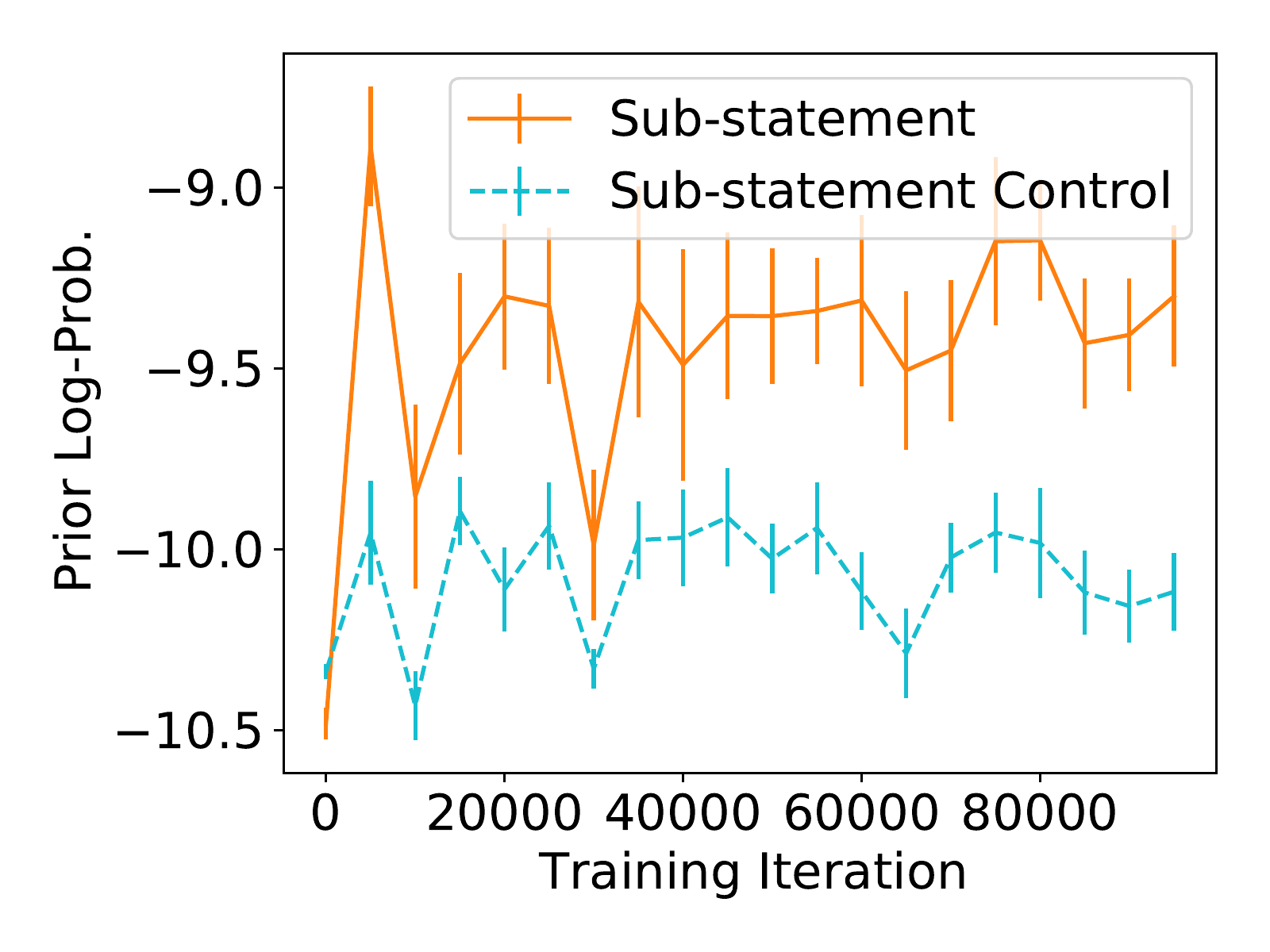} \label{fig:prior}}}
    \subfloat[\centering Super-statement->Sub-statement]{{\includegraphics[width=5cm]{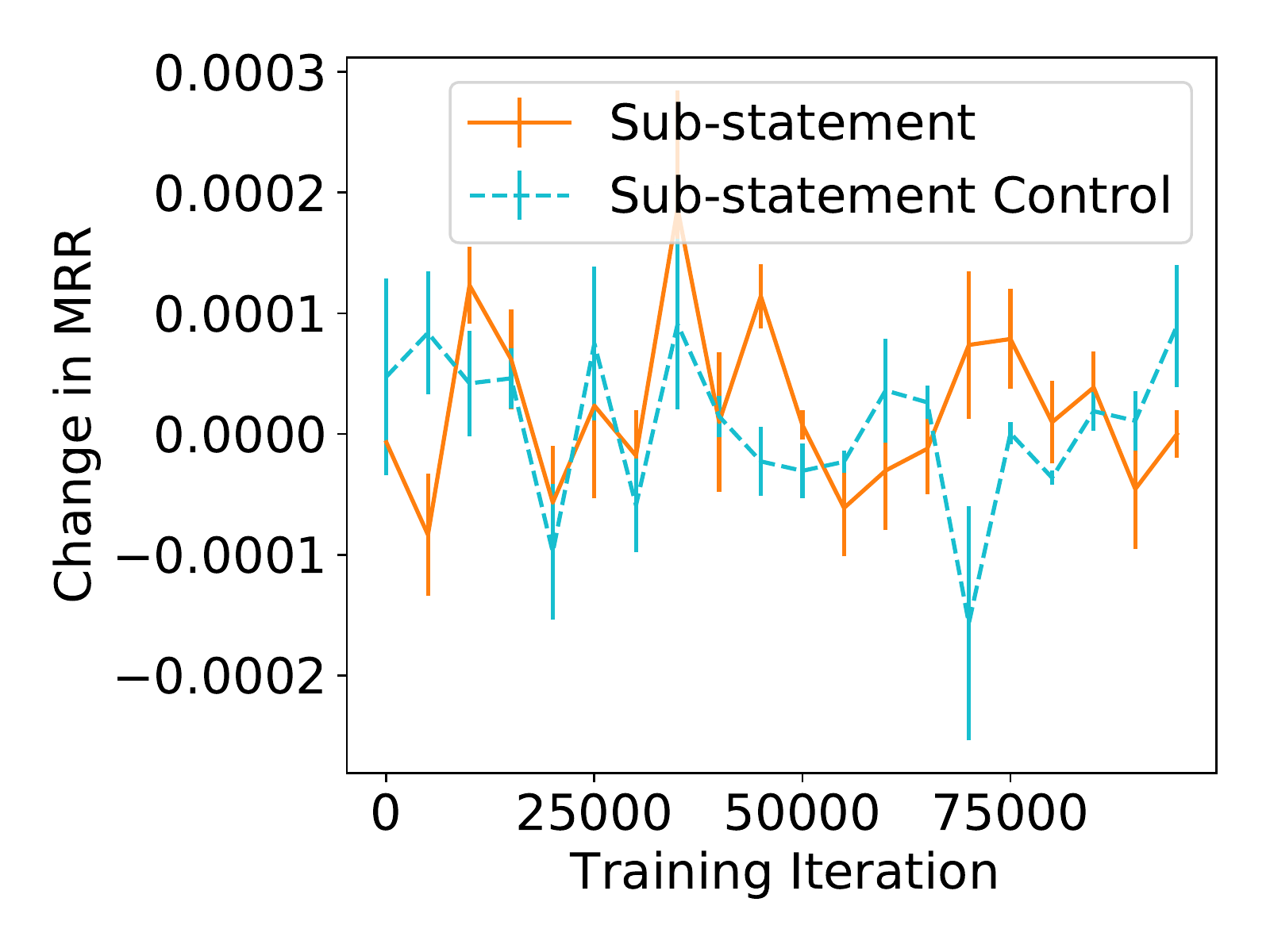} \label{fig:mrr}}}
    \\
    \subfloat[\centering Super-statement->Sub-statement]{{\includegraphics[width=5cm]{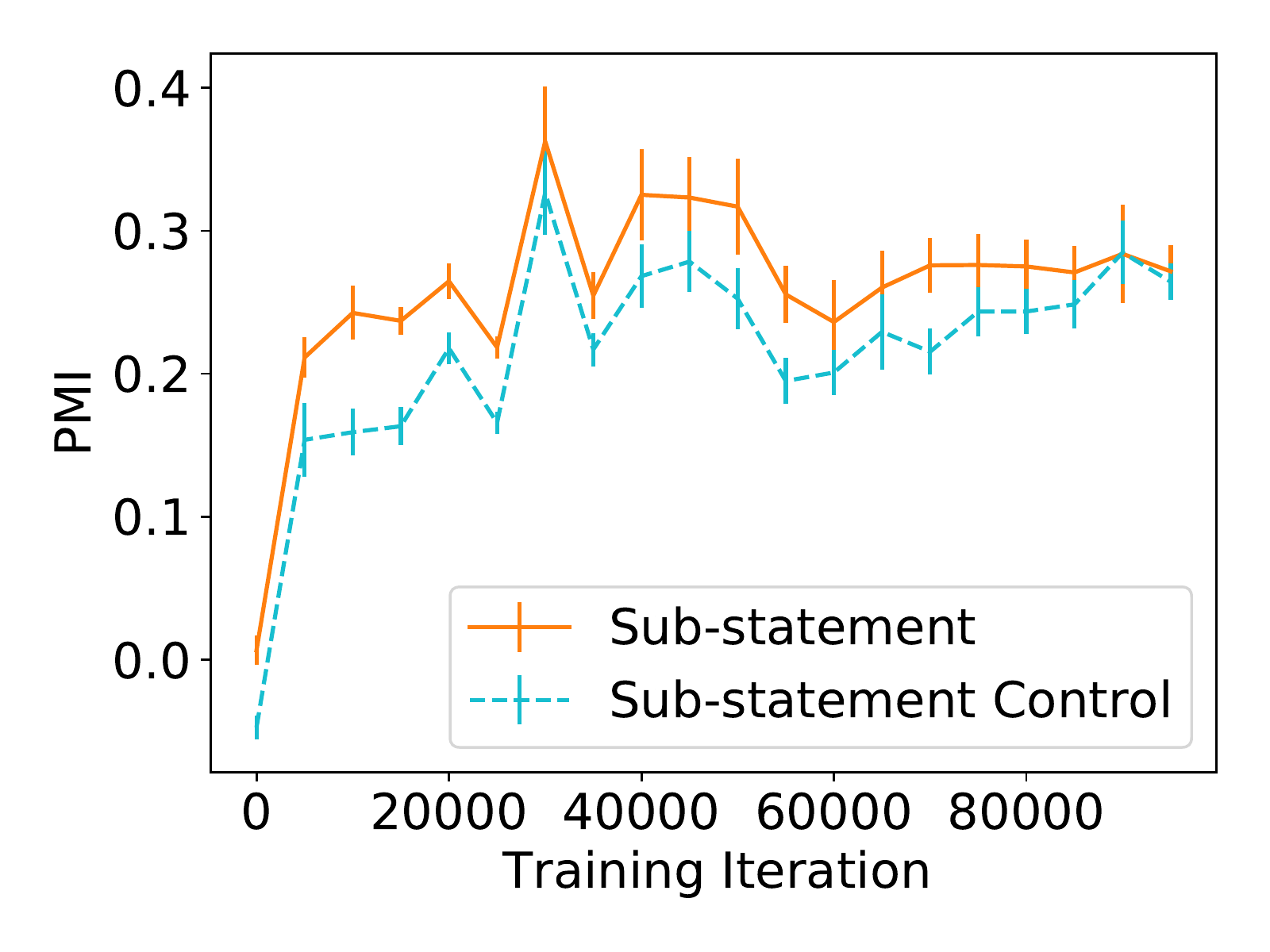} \label{fig:sub_statement_pmi}}}
    \subfloat[\centering Super-statement->Super-statement]{{\includegraphics[width=5cm]{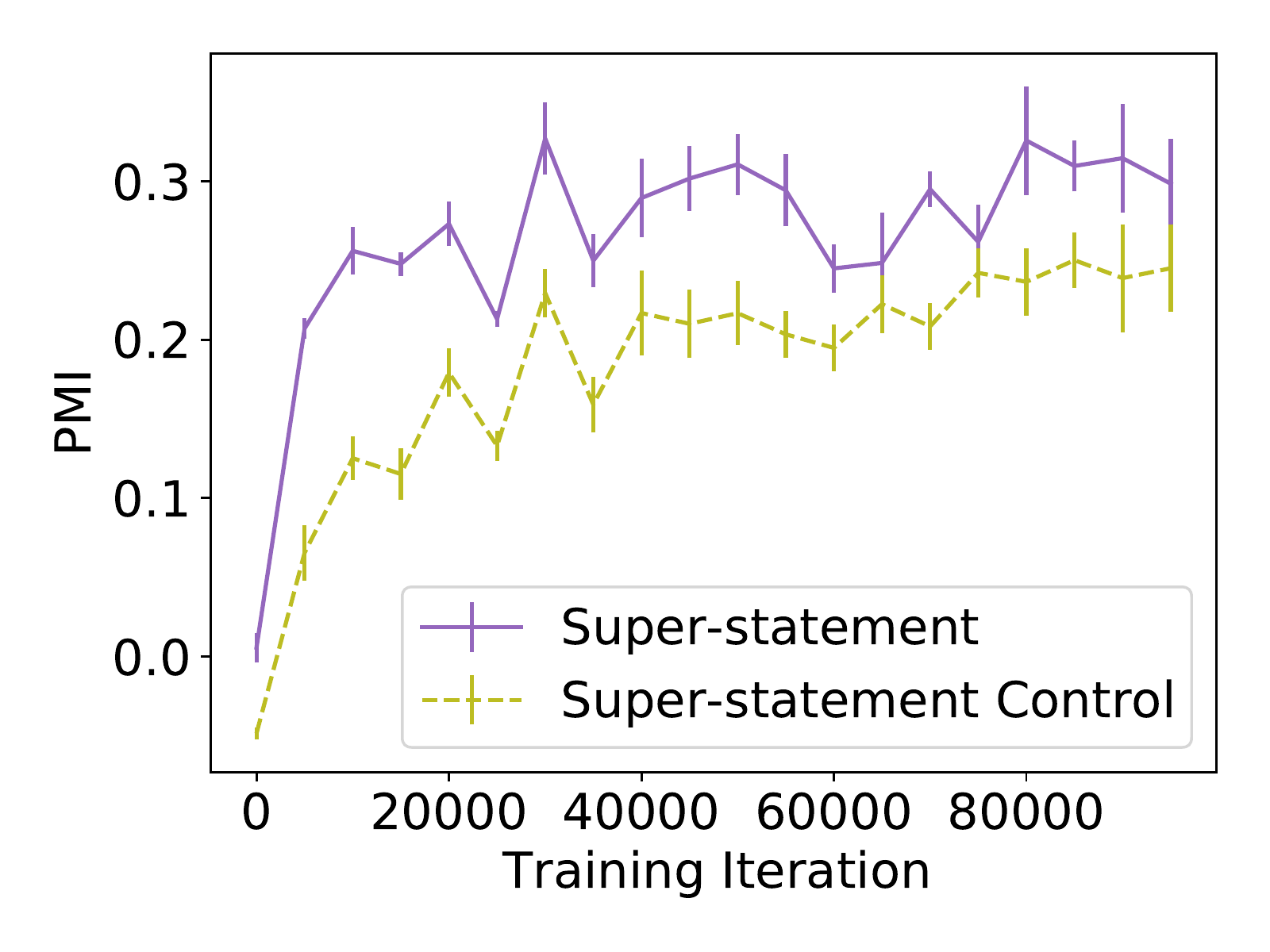} 
    \label{fig:super_statement_pmi}}}
    \subfloat[\centering Super-statement->Class-relation]{{\includegraphics[width=5cm]{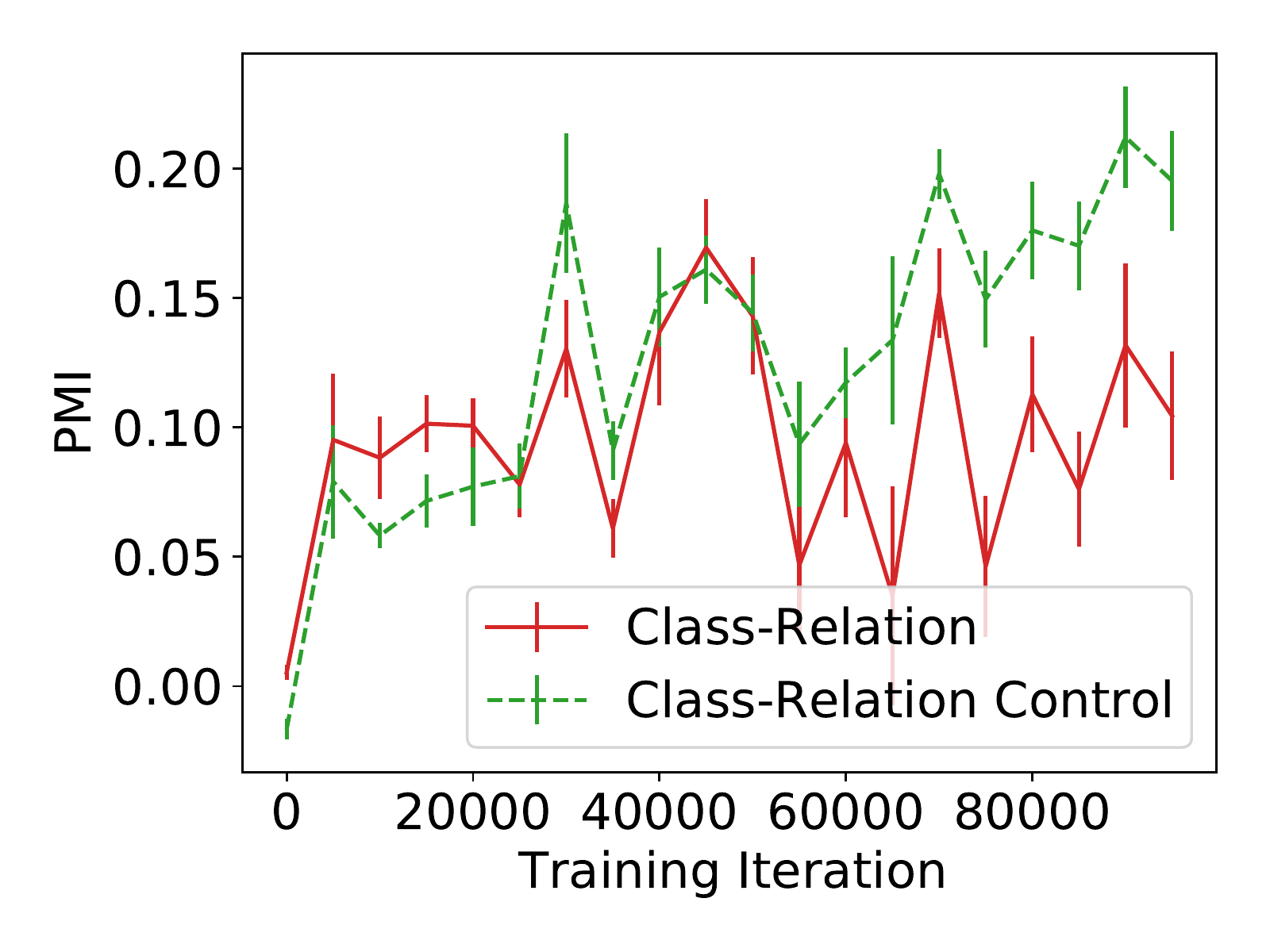} \label{fig:relation_pmi}}}
    \caption{Evaluations of BERT's knowledge and generalization across pre-training iterations. (a) shows the prior log-probability of sub-statements over time. (c-e) are the PMI for injecting supporting super-statements into the minibatch and evaluating on the labelled knowledge type. (b) is the change in MRR specifically for evaluating on sub-statements.}
    \label{fig:intervention}
\end{figure*}

\section{Results}
\label{sec:results}

\subsection{Fine-tuning}
\label{sec:finetuning}

We first run \citet{talmor-2020-leap}'s original fine-tuning evaluation on our pre-training checkpoints in order to validate our trained model and contrast fine-tuned evaluations with our pre-training interventions.

The explicit reasoning test requires classifying a sub-statement as true or false given the supporting super-statement and class-relation. The implicit reasoning test requires classifying a sub-statement given only the super-statement (and thus requires reasoning over implicit knowledge of the relation). Both test sets consist of disjoint entities from the training set. We fine-tune for 4 epochs following Talmor et al. and otherwise use default hyperparameters. The final implicit reasoning accuracy of our model is 0.89, slightly higher than \citet{talmor-2020-leap} report for RoBERTa-large.

We find performance increases log-linearly with pre-training iterations in the implicit reasoning test, but interestingly performance of the explicit reasoning evaluation peaks at just 15\% of pre-training (Figure \ref{fig:motivation}).  Numerical results are presented in Appendix \ref{sec:motivation-results}.

\subsection{Pre-training Interventions}
\label{sec:pretraining-interventions}

Figure \ref{fig:prior} shows the prior log-probability for our BERT model predicting sub-statement predicates during pre-training. The fact that the difference between the correct and control predicates increases during pre-training suggests knowledge of the sub-statement is acquired by BERT. Interestingly, probability of the correct predicate initially peaks at just 50k pre-training steps.

In \Cref{fig:mrr,fig:sub_statement_pmi,fig:super_statement_pmi,fig:relation_pmi}, we visualize the results of our pre-training interventions. At each pre-training checkpoint, we update BERT on a minibatch with injected super-statements and then evaluate on predicting the predicate of the labelled knowledge type. For \Cref{fig:sub_statement_pmi,fig:super_statement_pmi,fig:relation_pmi}, we consider PMI; i.e., how does updating on a super-statement affect the likelihood of supported knowledge?

When BERT is updated on a pre-training minibatch containing a super-statement, this unsurprisingly increases the probability of the super-statements predicate (\Cref{fig:super_statement_pmi}).

However, the PMI of the correct sub-statement predicate is the same as for the control predicate during the final iterations of pre-training (\Cref{fig:sub_statement_pmi}). Even more pronounced, the PMI of the class-relation control predicate is higher than the correct predicate during the entire second half of pre-training (\Cref{fig:relation_pmi}).

In other words, updating on statements such as ``birds can \mask{}'' $\rightarrow{}$ ``fly'' increases the BERT's estimate that ``robins are \mask{}'' $\rightarrow{}$ ``fish'' more than ``birds.'' We find a similar pattern in the reverse case: updating on class-relations improves BERT's knowledge of super-statements and sub-statements less than the control baselines.

If knowledge is supported by the appropriate inference rules, we would expect that updating on a premise would improve the knowledge of inferences. Furthermore, if this reasoning was being induced, we would expect this generalization from premise to inference to improve over time. And yet, we find the opposite to be true. It follows that the studied inferences are not inferred by BERT using the rule of downward monotonicity.

For \Cref{fig:mrr}, we consider the change in MRR of the sub-event predicate after updating on a minibatch containing the super-statement. In this case, the difference between predicting the correct and control predicate seems indiscernible across pre-training checkpoints.

\section{Conclusion}
\label{sec:conclusion}

We show that BERT does not acquire commonsense knowledge from premises and learned inferences. This highlights limitations of scaled pre-training and suggests developments in commonsense knowledge acquisition may require explicit reasoning mechanisms.



\bibliography{anthology,custom}
\bibliographystyle{acl_natbib}

\appendix

\section{BERT Hyperparameters}
\label{sec:bert-hyperparameters}

We follow the BERT-base architecture (12 layers, 12 attention heads, hidden size of 768) and train with the Adam optimizer. We use a sequence length of 128 throughout pre-training, and MegatronLM pre-processing. Training takes four days on eight V100 GPUs.

\newpage

\section{\lot{} Fine-tuning Results}
\label{sec:motivation-results}

\begin{table}[ht]
\begin{tabular}{ccc}
\toprule
Iteration & Implicit & Explicit \\
\midrule
0      & 0.507 & 0.493 \\
5000   & 0.507 & 0.493 \\
10000  & 0.490 & 0.490 \\
15000  & 0.571 & 0.621 \\
20000  & 0.625 & 0.636 \\
30000  & 0.710 & 0.763 \\
40000  & 0.798 & 0.900 \\
50000  & 0.814 & 0.965 \\
100000 & 0.838 & 0.971 \\
150000 & 0.860 & 0.992 \\
200000 & 0.843 & 0.953 \\
250000 & 0.855 & 0.973 \\
300000 & 0.870 & 0.958 \\
350000 & 0.863 & 0.978 \\
400000 & 0.850 & 0.931 \\
450000 & 0.867 & 0.937 \\
500000 & 0.859 & 0.933 \\
550000 & 0.874 & 0.951 \\
600000 & 0.867 & 0.943 \\
650000 & 0.880 & 0.931 \\
700000 & 0.877 & 0.937 \\
750000 & 0.874 & 0.929 \\
800000 & 0.872 & 0.949 \\
850000 & 0.877 & 0.979 \\
900000 & 0.875 & 0.967 \\
950000 & 0.894 & 0.945 \\
\bottomrule
\end{tabular}
\end{table}



\end{document}